\pdfoutput=1

\documentclass[11pt]{article}

\usepackage[]{acl}

\usepackage{times}
\usepackage{latexsym}

\usepackage[T1]{fontenc}

\usepackage[utf8]{inputenc}

\usepackage{microtype}

\usepackage{inconsolata}
\usepackage{multirow}
\usepackage{tabularx}
\usepackage{booktabs}
\usepackage{pifont}
\usepackage{xspace}
\usepackage{cleveref}
\usepackage{subfig}
\usepackage{subcaption}
\usepackage{lipsum}
\usepackage{hyperref}
\usepackage{url}
\usepackage{pdfpages}
\usepackage{soul}
\usepackage{fontawesome5}

\newcommand*{\affaddr}[1]{#1} 
\newcommand*{\affmark}[1][*]{\textsuperscript{#1}}
\newcommand*{\email}[1]{\texttt{#1}}

\usepackage{graphicx}
\usepackage{array}
\usepackage[table,xcdraw,dvipsnames, svgnames]{xcolor}
\usepackage{booktabs}
\usepackage{adjustbox}
\usepackage{enumitem}

\definecolor{FCE8E8}{HTML}{FCE8E8} 
\definecolor{E7F3E6}{HTML}{E7F3E6} 

\usepackage{highlight}

\newcolumntype{L}[1]{>{\raggedright\let\newline\\\arraybackslash\hspace{0pt}}p{#1}}
\newcolumntype{M}[1]{>{\raggedright\let\newline\\\arraybackslash\hspace{0pt}}m{#1}}

\definecolor{darkgreen}{rgb}{0.0, 0.4, 0.13}
\definecolor{lightlightblue}{rgb}{0.9, 0.95, 1.0}

\title{Leveraging Hierarchical Organization for Medical Multi-document Summarization}

\author{%
Yi-Li Hsu\affmark[1,2], Katelyn X. Mei \affmark[2],  Lucy Lu Wang\affmark[2,3]\\
\affaddr{\affmark[1]National Tsing Hua
University} \\
\affaddr{\affmark[2]University of Washington}\\ 
\affaddr{\affmark[3]Allen Institute for AI}\\
\email{yiligml@gapp.nthu.edu.tw; \{yilihsu, kmei, lucylw\}@uw.edu}
\\
\href{https://github.com/yilihsu/Hierarchical-MDS}{\faGithub\ \texttt{github.com/yilihsu/Hierarchical-MDS}}
}

\begin{document}
\maketitle
\begin{abstract}
Medical multi-document summarization (MDS) is a complex task that requires effectively managing cross-document relationships. This paper investigates whether incorporating hierarchical structures in the inputs of MDS can improve a model's ability to organize and contextualize information across documents compared to traditional flat summarization methods. We investigate two ways of incorporating hierarchical organization across three large language models (LLMs), and conduct comprehensive evaluations of the resulting summaries using automated metrics, model-based metrics, and domain expert evaluation of preference, understandability, clarity, complexity, relevance, coverage, factuality, and coherence. Our results show that human experts prefer model-generated summaries over human-written summaries. Hierarchical approaches generally preserve factuality, coverage, and coherence of information, while also increasing human preference for summaries. Additionally, we examine whether simulated judgments from GPT-4 align with human judgments, finding higher agreement along more objective evaluation facets. Our findings demonstrate that  hierarchical structures can improve the clarity of medical summaries generated by models while maintaining content coverage, providing a practical way to improve human preference for generated summaries.
\end{abstract}

\section{Introduction}
\begin{figure}[!ht]
    \centering
    \includegraphics[width=0.46\textwidth]{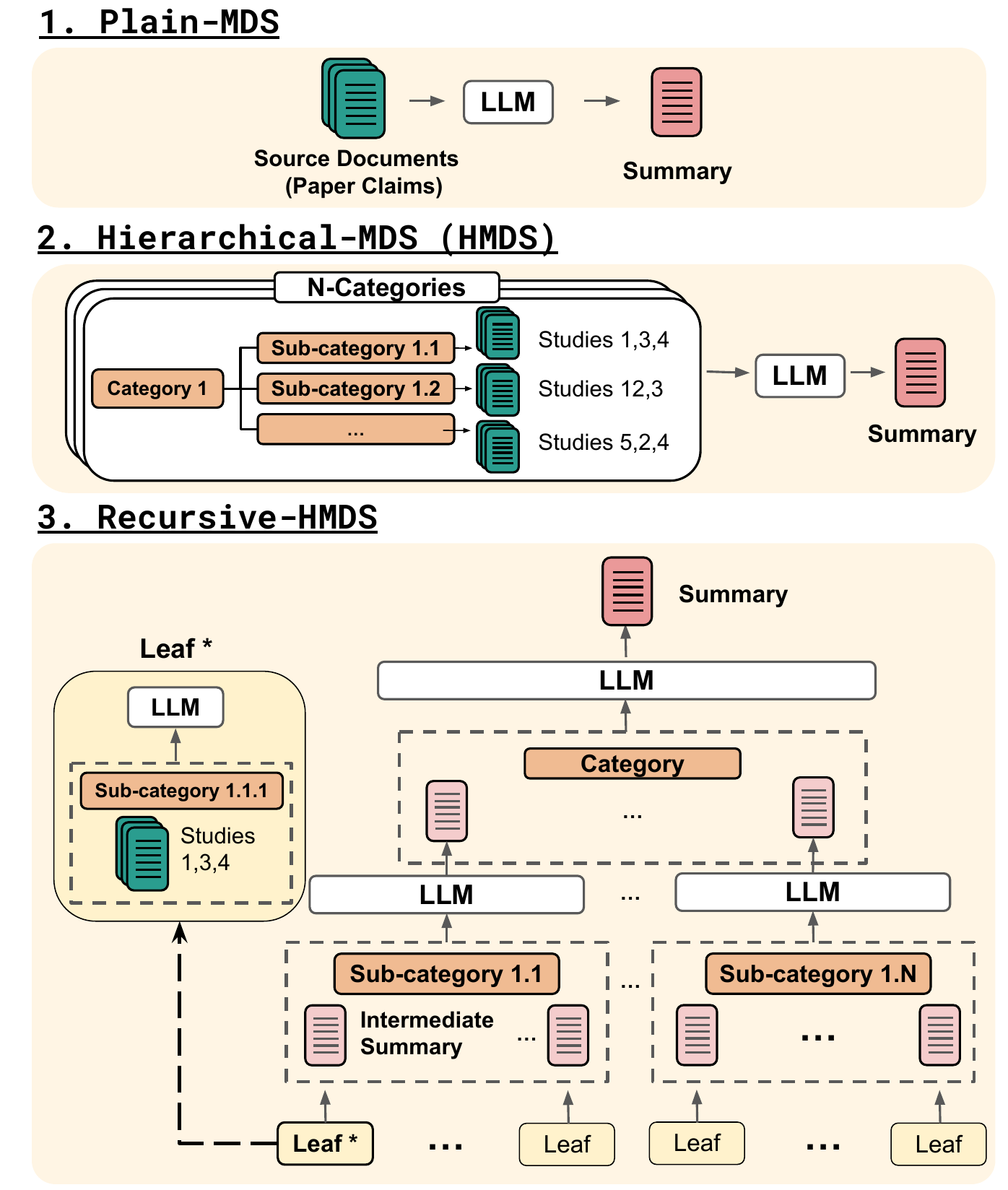}
    \caption{MDS settings we investigate: (1) Plain-MDS generates a summary by concatenating all source documents as input; (2) Hierarchical-MDS (HMDS) organizes source documents into categories before summarization; (3) Recursive-HMDS generates intermediate summaries for sub-categories first before iteratively producing the final summary.
    }
    \label{fig:main_diagram}
\end{figure}



Systematic reviews---aggregating and evaluating knowledge from multiple research efforts---is critical for the medical domain as it can inform clinical decision-making \cite{doi:10.1177/014107680309600304}.
Despite the importance of reviews for justifying clinical decisions, the average time from review initiation to publication is a long 67 weeks (15.5 months) \cite{Borah2017AnalysisOT, Luo2024PotentialRO, MICHELSON2019100443}. 
This delay 
prohibits a timely dissemination of knowledge given the rapid expansion of scientific research. 
Advancements in automated multi-document summarization (MDS) systems have shown significant promise in addressing these challenges \cite{zhang2024closing, Adams2021WhatsIA,Li2023SummarizingMD}. 
Researchers have adopted various approaches from data curation of high-quality human-written summaries to leveraging large language models 
 (LLMs) for information synthesis of research papers \cite{Nagar2024uMedSumAU, deyoung2021ms2, Wallace2020GeneratingN, Nair2023GeneratingMS}. 
Notably, recent studies demonstrate significant progress in retrieval-augmented LLMs, which enhance the process of answering scientific queries by identifying relevant passages and synthesizing comprehensive responses \cite{asai2024openscholar, lala2023paperqa}. Meanwhile, researchers also have developed techniques to extract structured data from medical literature \cite{wang2025foundation}.
 With the introduction of long-document transformers and promptable LLMs with long context windows, researchers have also begun investigating how hierarchical structure can be asserted in the inputs to these models. 
Hierarchical structures align with the human information-processing system \cite{CRAIK20019593} and mirror the way individuals perceive, interpret, and summarize complex information \cite{hasson2015hierarchical}. 
Interestingly, \citet{Wang2024DisentanglingII} find that adding hierarchical structure allows both humans and models to generate more semantically rich output representations compared to flat summarization approaches.
Indeed, \citet{hsu-etal-2024-chime} recently introduced an LLM-based framework (CHIME) that automatically extracts and organizes the main claims made in multiple cited medical papers within a review into hierarchies; their work argues that hierarchical organization could support researchers conducting systematic literature reviews. 

Inspired by these findings, we investigate whether incorporating hierarchical organization can enhance the quality of LLM-generated summaries in 
medical MDS. 
Leveraging the CHIME dataset \citep{hsu-etal-2024-chime}, we investigate various strategies for incorporating hierarchical organization in MDS, evaluating three models---GPT-4, Claude 3, and Mistral-7B---across three settings: (i) Plain-MDS (concatenating inputs), (ii) Hierarchical-MDS (HMDS), which concatenates the inputs along with category labels based on hierarchical organization, and (iii) Recursive HMDS, which recursively generates intermediary summaries before combining them into a final summary.
We evaluate summaries using a wide range of automated, model-based, and prompt-based LLM metrics and by engaging two groups of experts 
to annotate results along evaluation dimensions identified in prior work such as factuality \cite{10.1145/3545176}, coverage \cite{10.1145/1233912.1233913, Bao2022DocAsRefAE}, coherence \cite{10.1145/3545176}, overall preference, clarity, understandability, complexity \cite{10.1093/pnasnexus/pgae387}, and relevance \cite{Peyrard2018AST}. 
Since scaling expert feedback becomes increasingly challenging as the volume of labeled samples grows, we also explore the use of LLMs as simulated 
experts, following the approach of \citet{zhang2024closing}. 

We summarize our contributions as follows:

\begin{itemize}[itemsep=0pt, topsep=1pt, leftmargin=10pt]
\item We investigate the impact of incorporating hierarchical organization in medical MDS, comparing single-pass and recursive hierarchical approaches against a concatenation baseline (\S \ref{sec:methods});
\item We design three annotation tasks to collect expert judgments of summary quality along 7 dimensions (\S\ref{sec:annotation}). Compared to concatenated input, 
hierarchical methods improve expert-perceived clarity, understandability, and preference, while also reducing complexity, although no significant improvement is found by NLP metrics (\S\ref{sec:metrics}).

\item We explore the agreement between automated metrics, human judgments, and GPT-4-simulated judgments. We find that automated factuality metrics fail to capture experts assessed factuality, while GPT-4 aligns well with expert judgments on factuality and complexity. However, discrepancies exist in more subjective areas (e.g. relevance, coherence, and clarity), highlighting challenges in simulating expert judgments (\S \ref{sec:gpt4_evaluator}). 
\end{itemize}

\section{Related Works}
\paragraph{Medical Multi-document Summarization (MDS)}
Medical MDS has seen significant advances, particularly in systems designed to assist medical decisions~\cite{liang-etal-2019-novel-system, ahmed2023natural, reyes2015clinical}
and documentation of clinical notes~\cite{nair-etal-2024-dera, yim-yetisgen-2021-towards, krishna-etal-2021-generating, adams-etal-2021-whats}. The field has recently seen a shift from using pre-trained language models (PLMs) such as BioBERT \cite{10.1093/bioinformatics/btz682} and ClinicalBERT \cite{Huang2019ClinicalBERTMC} to leveraging LLMs to synthesize findings from multiple research papers and summarize clinical conversations \cite{zhang2024closing, woo2024synthetic, Godbole2024LeveragingLL}. 
A notable challenge is the difficulty of accurately synthesizing evidence from multiple clinical trials and studies 
\cite{shaib-etal-2023-summarizing}.
Our study aims to explore whether incorporating hierarchical organization can improve factual accuracy for medical MDS.

\paragraph{Evaluating Medical MDS}
 Over the years, researchers have developed a variety of automated metrics focusing on evaluating factuality and coverage of generated summaries~\cite{amrfact2024, fabbri2020summeval, huang2020ffci, chern2023factool, soleimani2023nonfacts}. 
Despite these advancements, a persistent gap remains: automated evaluation metrics do not always correlate with human judgments \cite{sottana-etal-2023-evaluation, wang2023automated, howcroft2024survey, 10.1145/3641289}. 
Our paper further investigates this evaluation gap, by studying correlation between automated metrics, human preferences, as well as simulated GPT-4 evaluations in hierarchical medical MDS.

\paragraph{Hierarchical Structures in MDS}
The hierarchical structure of documents has been utilized to enhance search, reading comprehension, and knowledge acquisition by offering a clear and informative overview of the content \cite{Taylor1984TheEO, Guthrie1991RolesOD, Li2023SummarizingMD}.  Previous research has incorporated hierarchical information into MDS via modifying model architectures such as LSTMs \cite{Fabbri2019MultiNewsAL}, encoder-decoder networks \cite{Liu2019HierarchicalTF, Karn2019AHD}, and split-then-summarize frameworks \cite{Zhang2021SummNAM, ravaut-etal-2024-context}. Other studies generate categorical hierarchies for 
source studies to assist literature reviews \cite{Zhu2023HierarchicalCG, hsu-etal-2024-chime}. To our knowledge, no studies have yet explored how hierarchical information can be better incorporated to assist LLMs in generating medical MDS, particularly for literature reviews. Our paper investigates how utilizing hierarchical organization
can impact the quality of the output, potentially offering 
improvements in the summarization process for research applications.

\section{Dataset \& Model Settings}
\label{sec:methods}

\subsection{Dataset}

We conduct experiments on the CHIME dataset \cite{hsu-etal-2024-chime}, which consists of expert-written review abstracts from the Cochrane Database of Systematic Reviews\footnote{https://www.cochranelibrary.com/cdsr/reviews}, claims extracted from studies included in review abstracts, and LLM-generated hierarchical organization for these claims annotated by experts. Each review in CHIME includes an average of 24.7 (min=15, max=50) corresponding studies, and one-sentence claims based on these studies generated by Claude-2 and validated by the authors. 
The hierarchies are defined as tree structures, where nodes and sub-nodes represent topical and sub-topical categories, and each leaf node is linked to the studies in that category. All hierarchies are
multi-level, with a mean hierarchy depth of 2.5, and maximum depth of 5. Claims are organized into both LLM-generated and expert-curated hierarchies. While we experiment with both, our focus is on LLM-generated hierarchies, as they more accurately reflect the capabilities of models in an end-to-end setting.
We randomly sample 30 topics from CHIME and 
use claims as input documents instead of paper abstracts because the automatic metric scores for claims are higher than those for paper abstracts
 (details in Appendix~\ref{sec:condense}).
For the expert-written review abstracts, we retain the ``Main Results'' and ``Author's Conclusion'' sections from each review abstract to compare with LLM-generated summaries regarding the qualities of summarized main findings.   
In the following experiments, we explore different ways of integrating hierarchical organization into MDS and evaluate their impact on summary quality.

\subsection{Summarization Settings}
\label{sec:sum_setting}
We conduct experiments with three different language models: GPT-4  \cite{achiam2023gpt}, Claude-3 \cite{claude}, and Mistral-7B \cite{jiang2023mistral}.\footnote{We employ GPT-4 (gpt-4-0613), Claude 3 (claude-3-opus-20240229), and Mistral-7B (Mistral-7B-Instruct-v0.2). We set the output truncation to 1024 tokens. The temperature and top-p sampling settings were kept at the models' default values to ensure optimal performance.} 
We experiment with three summarization settings: (i) Plain MDS without Hierarchies; (ii) Hierarchical MDS (HMDS); and (iii) Recursive-HMDS (Figure~\ref{fig:main_diagram}), 
detailed below along with the corresponding prompt in Appendix Fig.~\ref{fig:prompt}.

\vspace{-1mm}
\paragraph{Plain MDS} This is our baseline setting, where the input is formed by concatenating all claims without injecting any hierarchical information. 

\vspace{-1mm}
\paragraph{Hierarchical MDS}
 We incorporate hierarchies in the input by creating a nested list that includes claims and their respective categories extracted from CHIME. 
We experiment with LLM-generated hierarchies and expert-curated hierarchies from CHIME to organize the inputs.

\vspace{-1mm}
\paragraph{Recursive-HMDS}
This setting leverages the hierarchical structure through a recursive summarization approach. Unlike typical MDS methods that process all source documents in one pass, this approach recursively combines summaries from lower-level summaries. In other words, we first summarize all claims in each \textit{leaf category}, the most granular level in our hierarchy. The summaries of leaf categories are used as inputs to produce intermediary summaries of the hierarchy, and this is repeated until we reach the root of the tree. This approach mirrors the bottom-up strategy used by human experts when conducting systematic reviews, where information is first organized at a granular level and then consolidated into higher-level summaries. In this setting, we use expert-curated hierarchies from CHIME to assess how well the curation of hierarchical organization enhances summary qualities.







\section{Automated Evaluation Metrics}
\label{sec:metrics}
We employ a comprehensive suite of automated and model-based metrics to assess the quality of text generated by language models. These include:

\begin{itemize}[noitemsep, topsep=1pt, leftmargin=10pt]
\item ROUGE \cite{lin2004rouge}: We report ROUGE-L Recall for evaluating the overlap of n-grams between the summaries and source documents.
\item BERT-Score \cite{DBLP:journals/corr/abs-1904-09675}: evaluates contextual semantic similarity between generated texts and source documents with BERT model.
\item Pyramid Score \cite{zhang2021finding}: assesses coverage by calculating how much of the semantic content units (SCUs) from source documents are entailed by the generated summary.
\item Reversed-Pyramid Score: assess factuality by measuring how the SCUs from generated summaries are entailed by the source documents.
\item FIZZ \cite{yang-etal-2024-fizz}: assess factual consistency by measuring entailment between atomic facts in the generated summaries and the source document. We report two variants: (i) min-max FIZZ from original paper, which captures the worst-case entailment gaps, and (ii) average FIZZ, which reflects overall factual consistency by averaging the entailment scores of atomic facts against all sentences in the summary. 

\item LLM prompt-based evaluation \cite{guo-etal-2024-appls, gao2023human}: 
we prompt GPT-4 using the template specifically designed for summarization tasks. The template guides GPT-4 to assess summaries based on informativeness, simplification, coherence, and faithfulness. We conduct both reference-based and reference-free methods provided in \cite{guo-etal-2024-appls} with GPT-4 needed to provide an explanation for each score. 
\end{itemize}

\noindent  We report the results in Appendix Table~\ref{tab:autoresults}, and more detailed descriptions of these evaluation methods
 are provided in Appendix \ref{sec:app_metrics}. We discuss how these metrics correlate with human assessments in \S\ref{sec:metrics_corr_human}.

\section{Expert Annotation Tasks}
\label{sec:annotation}
We assess overall subjective preferences among the generated summaries through pairwise comparisons, evaluating dimensions including overall preference, clarity, understandability, complexity \cite{10.1093/pnasnexus/pgae387}, and relevance \cite{Peyrard2018AST} (Tasks 1 and 2). Additionally, Task 3 is dedicated to evaluating key objective dimensions widely recognized in the NLP community, including factuality \cite{10.1145/3545176}, coverage \cite{10.1145/1233912.1233913, Bao2022DocAsRefAE} and coherence \cite{10.1145/3545176}. Notably, our task assesses factuality and coverage against source paper abstracts, an approach seldom adopted in previous MDS studies that typically use review abstracts as references. 


\paragraph{Data Sampling:} 
Given the time- and cost-intensive nature of quality annotation, we annotated 30 topics for Tasks 1 and 2, and 27 topics for Task 3. To ensure our sample size was sufficient in evaluating statistical significance, we employ a power analysis to determine the required number of annotations. 



\paragraph{Annotator Background} 
\label{para:annotation_details}  Annotations are completed by domain experts with experience reading medical literature. We recruited two separate groups of domain experts through Upwork to annotate each task: four annotators for Tasks 1 and 2 and two for Task 3. All annotators have completed MS/PhD in Biomedicine, Biology, Microbiology, or Neuroscience. Each annotator was required to complete a pilot batch to verify annotation quality.
\footnote{Agreement is assessed in the pilot batch for Task 3 (coverage and factuality)
We compute pairwise Cohen's weighted kappa with quadratic weighting, achieving scores of 0.69 for coverage and 0.52 for factuality.} Annotators were compensated based on the total number of hours spent on the tasks, with 55 hours worked across 4 annotators in Tasks 1 and 2 and 50 hours worked across 2 annotators in Task 3. Annotators were paid \$25 per hour. 





\begin{table*}[ht]
\centering
\begin{adjustbox}{max width=\textwidth}
\begin{tabular}{lllllll}
\toprule
\textbf{Setting} & \textbf{Model} & \textbf{Overall} $\uparrow$ & \textbf{-Complex} $\uparrow$  & \textbf{Relevance} $\uparrow$ & \textbf{Clarity} $\uparrow$ & \textbf{Understand} $\uparrow$ \\
\midrule

Review Abstract & - & \midpointgradientcellog{0.2}{0.1}{0.8}{0}{neg}{pos}{\opacity}{0} & \midpointgradientcellog{0.044}{0.1}{0.8}{0}{neg}{pos}{\opacity}{0} & \midpointgradientcellog{0.578}{0.1}{0.8}{0}{neg}{pos}{\opacity}{0} & \midpointgradientcellog{0.089}{0.1}{0.8}{0}{neg}{pos}{\opacity}{0} & \midpointgradientcellog{0.089}{0.1}{0.8}{0}{neg}{pos}{\opacity}{0} \\
\midrule 

Plain MDS & \multirow{2}{*}{GPT-4} & \midpointgradientcellog{0.537}{0.1}{0.8}{0}{neg}{pos}{\opacity}{0} & \midpointgradientcellog{0.463}{0.1}{0.8}{0}{neg}{pos}{\opacity}{0} & \midpointgradientcellog{0.317}{0.1}{0.8}{0}{neg}{pos}{\opacity}{0} & \midpointgradientcellog{0.571}{0.1}{0.8}{0}{neg}{pos}{\opacity}{0} & \midpointgradientcellog{0.512}{0.1}{0.8}{0}{neg}{pos}{\opacity}{0} \\
HMDS w/ LLM-Hierarchy &   & \midpointgradientcellog{0.568}{0.1}{0.8}{0}{neg}{pos}{\opacity}{0} & \midpointgradientcellog{0.568}{0.1}{0.8}{0}{neg}{pos}{\opacity}{0} & \midpointgradientcellog{0.5}{0.1}{0.8}{0}{neg}{pos}{\opacity}{0} & \midpointgradientcellog{0.523}{0.1}{0.8}{0}{neg}{pos}{\opacity}{0} & \midpointgradientcellog{0.523}{0.1}{0.8}{0}{neg}{pos}{\opacity}{0} \\
\midrule

Plain MDS & \multirow{2}{*}{Claude 3} & \midpointgradientcellog{0.814}{0.1}{0.8}{0}{neg}{pos}{\opacity}{0} & \midpointgradientcellog{0.814}{0.1}{0.8}{0}{neg}{pos}{\opacity}{0} & \midpointgradientcellog{0.163}{0.1}{0.8}{0}{neg}{pos}{\opacity}{0} & \midpointgradientcellog{0.814}{0.1}{0.8}{0}{neg}{pos}{\opacity}{0} & \midpointgradientcellog{0.791}{0.1}{0.8}{0}{neg}{pos}{\opacity}{0} \\
HMDS w/ LLM-Hierarchy &   & \midpointgradientcellog{0.578}{0.1}{0.8}{0}{neg}{pos}{\opacity}{0} & \midpointgradientcellog{0.889}{0.1}{0.8}{0}{neg}{pos}{\opacity}{0} & \midpointgradientcellog{0.111}{0.1}{0.8}{0}{neg}{pos}{\opacity}{0} & \midpointgradientcellog{0.756}{0.1}{0.8}{0}{neg}{pos}{\opacity}{0} & \midpointgradientcellog{0.556}{0.1}{0.8}{0}{neg}{pos}{\opacity}{0} \\
\midrule

Plain MDS &  \multirow{4}{*}{Mistral-7B} & \midpointgradientcellog{0.362}{0.1}{0.8}{0}{neg}{pos}{\opacity}{0} & \midpointgradientcellog{0.34}{0.1}{0.8}{0}{neg}{pos}{\opacity}{0} & \midpointgradientcellog{0.404}{0.1}{0.8}{0}{neg}{pos}{\opacity}{0} & \midpointgradientcellog{0.298}{0.1}{0.8}{0}{neg}{pos}{\opacity}{0} & \midpointgradientcellog{0.213}{0.1}{0.8}{0}{neg}{pos}{\opacity}{0} \\
HMDS w/ LLM-Hierarchy & & \midpointgradientcellog{0.391}{0.1}{0.8}{0}{neg}{pos}{\opacity}{0} & \midpointgradientcellog{0.435}{0.1}{0.8}{0}{neg}{pos}{\opacity}{0} & \midpointgradientcellog{0.37}{0.1}{0.8}{0}{neg}{pos}{\opacity}{0} & \midpointgradientcellog{0.413}{0.1}{0.8}{0}{neg}{pos}{\opacity}{0} & \midpointgradientcellog{0.326}{0.1}{0.8}{0}{neg}{pos}{\opacity}{0} \\
HMDS w/ Expert-Hierarchy &  & \midpointgradientcellog{0.37}{0.1}{0.8}{0}{neg}{pos}{\opacity}{0} & \midpointgradientcellog{0.348}{0.1}{0.8}{0}{neg}{pos}{\opacity}{0} & \midpointgradientcellog{0.457}{0.1}{0.8}{0}{neg}{pos}{\opacity}{0} & \midpointgradientcellog{0.37}{0.1}{0.8}{0}{neg}{pos}{\opacity}{0} & \midpointgradientcellog{0.304}{0.1}{0.8}{0}{neg}{pos}{\opacity}{0} \\
Recursive-HMDS w/ Expert-Hierarchy & & \midpointgradientcellog{0.578}{0.1}{0.8}{0}{neg}{pos}{\opacity}{0} & \midpointgradientcellog{0.622}{0.1}{0.8}{0}{neg}{pos}{\opacity}{0} & \midpointgradientcellog{0.267}{0.1}{0.8}{0}{neg}{pos}{\opacity}{0} & \midpointgradientcellog{0.622}{0.1}{0.8}{0}{neg}{pos}{\opacity}{0} & \midpointgradientcellog{0.467}{0.1}{0.8}{0}{neg}{pos}{\opacity}{0} \\

\bottomrule
\end{tabular}
\end{adjustbox}
\caption{Win Rates (\%) through Random Pairwise Comparisons Across All Models and Settings (Task 1). Win rate for each setting is calculated as the percentage of comparisons where the setting was preferred over its counterpart.
Results indicate that model-generated summaries are generally preferred over review abstracts across all dimensions except for relevance, regardless of whether hierarchical organization is incorporated. 
}
\

\label{tab:comparison}
\end{table*}

\paragraph{Task 1: Random Pairwise Comparison across all models and settings} 
\label{sec:pairwise}
This task aims to evaluate the overall subjective quality of summaries by comparing pairs of summaries generated under different settings for the same topic. Drawing on prior work \cite{10.1093/pnasnexus/pgae387, oppenheimer2006consequences, 2daa416777ec457488c6fff0fd6b9f59, Peyrard2018AST}, we assess \textit{clarity} ("Which summary is clearer in its writing?"), \textit{complexity} ("Which summary is more complex in its writing?"),\footnote{We report 1-Complexity in all tables and figures to align the directions of all evaluation dimensions.} and \textit{understandability} ("Which summary do you understand better?"). To provide a more comprehensive evaluation of topic-centric MDS, we add \textit{relevance} ("Which summary is more relevant to the topic?") and \textit{overall preference} ("Which summary do you prefer overall?"). 
For each comparison, annotators are presented with two summaries from system A and system B and asked the above questions, with the response options for each being A, B, or about the same. To reduce bias, the order of systems is randomized and the settings or models used for the summaries are not disclosed to the annotators. The annotation interface is shown in Appendix Figure~\ref{fig:eval}.
We evaluate Plain MDS, HMDS with LLM-Hierarchy, and HMDS with Expert-Hierarchy, each implemented using GPT-4, Claude 3, and Mistral-7B; Recursive-HMDS with Expert-Hierarchy using Mistral; and human-written review abstracts as a human baseline. 
Each setting is randomly paired with another one for 5-6 times per comparison pairing. This results in comparing each setting for an average of 45 times (min=41, max=47). We report the average win rate of each setting in Table~\ref{tab:comparison}. 

\begin{figure*}[!ht]
    \centering
    \includegraphics[scale=0.73]{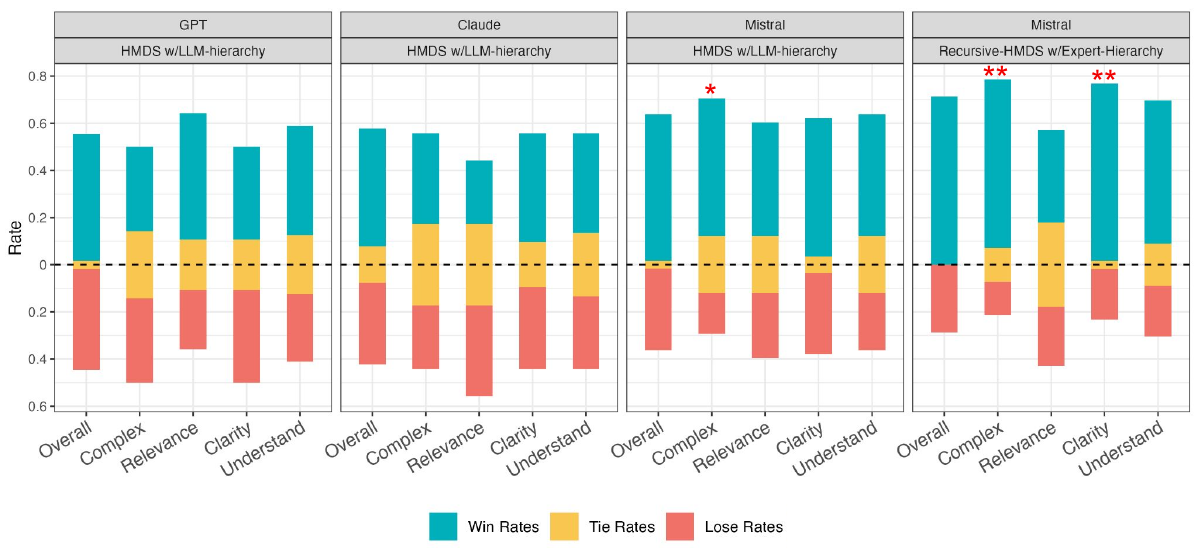} 
    \caption{Human pairwise evaluations for within-model comparisons (Task 2): LLM-generated summaries of hierarchical settings are compared to each model's baseline setting (Plain MDS). We report 1-Complexity to ensure alignment in the direction of all dimensions. The red star (\textcolor{red}{$\star$}) indicates the significance levels, with \textcolor{red}{$\star$$\star$} (p < 0.01) and \textcolor{red}{$\star$} (p < 0.05), as determined by McNemar's test (within-subject Chi-square test) based on the counts of wins and ties of each pair. 
    The results demonstrate that hierarchical setups can reduce complexity across Mistral-7B and Claude 3
    and improve overall preference, understandability, and clarity, particularly for Mistral-7B
    . 
    }
    \label{fig:wintie}
\end{figure*}

\paragraph{Task 2: Pairwise Comparison within Models: \textit{Hierarchical vs.~Non-hierarchical}}
To further explore the impact of hierarchical organization on MDS, we up-sample within-model comparisons from Task 1 for the Plain MDS versus single-pass hierarchical setting (HMDS) for all models. We also employed Recursive-HMDS using Mistral-7B for a more robust comparison on different hierarchical configurations. Each hierarchical setting with its non-hierarchical counterparts (Plain-MDS) was compared 30 times. Four annotators performed 297 pairwise comparisons. 
 We report the average win rate per system setting in Figure \ref{fig:wintie}). Given the identical annotation structure across Tasks 1 and 2, we employed the same group of annotators and present the inter-rater agreement in Table~\ref{tab:human_gpt_aggrement_task1}.



\paragraph{Task 3: Likert-Scale Assessment against Source Paper Abstracts} 
\label{sec:likert}

In this task, we assess objective qualities of summary quality.
Comparing each summary against source paper abstracts, annotators are asked to evaluate the \textit{coverage} (the summary covers the key claim made in the abstract), \textit{factuality} (the summary accurately reflects the meaning and details of the key claim in the abstract), and \textit{coherence} (the summary flows logically, with ideas connecting smoothly from one to the next without confusing jumps). Annotators rate each summary against 10 randomly selected source paper abstracts on coverage and factuality using a 5-point Likert scale, ranging from 1 (Disagree) to 5 (Agree). We compute the average Likert score based on the judgments from the 10 paper abstracts for each summary. Coherence is assessed once for the summary on the same Likert scale. We evaluate Plain MDS, HMDS with LLM-Hierarchy and Expert-Hierarchy, each implemented using GPT-4, Claude 3, and Mistral-7B; Recursive-HMDS with Expert-Hierarchy using Mistral; and human-written review abstract. 
Inter-rater agreement is 52\% for coverage and 69\% for factuality using Cohen's Kappa (quadratic weighted). 
Annotation interfaces are in Appendix Figures \ref{fig:eval-2} and \ref{fig:eval-3}.
    

\section{LLM-simulated Annotation}
\label{sec:simulation}

Scaling expert feedback becomes increasingly challenging as the volume of 
samples grows.
Therefore, we are motivated to explore the use of LLMs to simulate expert preferences 
in our annotations. We experiment with GPT-4 for simulating expert judgments, following the methodology outlined by \citet{zhang2024closing}, which prompts GPT-4 with the same questions that human annotators receive. To ensure a fair comparison, GPT-4 evaluate using the same set of summaries labeled by human experts. We conduct both zero-shot and 8-shot settings for pairwise comparison (Tasks 1 and 2), and a zero-shot setting for Likert-scale assessment (Task 3). The evaluation prompts are provided in Appendix Figures~\ref{fig:prompt_task1} and~\ref{fig:prompt_task2}.\footnote{We hypothesize that subjective dimensions assessed by Tasks 1 and 2 may benefit most from few-shot learning to enhance contextual understanding.} 
In the 8-shot setting, we sample eight expert annotations from the pilot batch, selecting two from each of the four annotators. 
We report inter-rater agreement 
in Table~\ref{tab:human_gpt_aggrement_task1} (Tasks 1 and 2) and Table~\ref{tab:likeret_combined} (Task 3). The simulated annotation results are reported in Table \ref{tab:comparison} (Task 1) and Appendix Table~\ref{fig:gpt-8shots} (Task 2).

\section{Results}
\label{sec:results}




\begin{table*}[ht]
\centering
\footnotesize
\renewcommand{\arraystretch}{0.6} 
\setlength{\tabcolsep}{4pt} 
\setlength{\extrarowheight}{-1pt}

\resizebox{1\textwidth}{!}{
\begin{tabular*}{\textwidth}{@{\extracolsep{\fill}}L{27mm}cM{10mm}cccccc}

\toprule
\multirow{2}{*}{\textbf{Setting}}  & \multirow{2}{*}{\textbf{Model}}  & \textbf{Avg Words} & \multicolumn{2}{c}{\textbf{Coverage}} & \multicolumn{2}{c}{\textbf{Factuality}} & \multicolumn{2}{c}{\textbf{Coherence}} \\
\cmidrule(lr){4-5} \cmidrule(lr){6-7} \cmidrule(lr){8-9}
 & &  & \textbf{Expert} & \textbf{GPT-4} & \textbf{Expert} & \textbf{GPT-4} & \textbf{Expert} & \textbf{GPT-4} \\
\midrule
Review Abstract & - & 379  & 3.07 ± 0.29 & 3.03 ± 0.21 & 2.30 ± 0.28 & 3.03 ± 0.21 & 3.73 ± 0.45 & 3.27 ± 0.24 \\
\midrule
Plain MDS & Mistral-7B & 515 & 3.86 ± 0.18 & 4.06 ± 0.20 & 3.18 ± 0.17 & 4.06 ± 0.20 & 4.17 ± 0.41 & 3.92 ± 0.23 \\
\midrule
HMDS w/ LLM-Hierarchy & GPT-4 & 301 & 3.74 ± 0.27 & 3.79 ± 0.19 & 3.09 ± 0.31 & 3.79 ± 0.19 & 4.44 ± 0.38 & 4.00 ± 0.24 \\
                           & Claude 3 & 195 & 3.60 ± 0.28 & 3.81 ± 0.11 & 3.08 ± 0.27 & 3.81 ± 0.11 & 4.40 ± 0.40 & 4.00 ± 0.00 \\
                           & Mistral-7B & 354 & 3.25 ± 0.28 & 3.51 ± 0.21 & 2.62 ± 0.28 & 3.51 ± 0.21 & 4.45 ± 0.31 & 3.64 ± 0.20 \\
\midrule
HMDS w/ Expert-Hierarchy & Mistral-7B & 301 & 3.81 ± 0.23 & 4.04 ± 0.20 & 3.27 ± 0.21 & 4.04 ± 0.20 & 3.50 ± 0.48 & 4.08 ± 0.15 \\
\midrule
Recursive-HMDS w/ Expert-Hierarchy & Mistral-7B & 252 & 3.51 ± 0.18 & 3.52 ± 0.16 & 2.70 ± 0.24 & 3.52 ± 0.16 & 4.05 ± 0.41 & 3.40 ± 0.16 \\

\bottomrule
\end{tabular*}
}
\caption{Mean (± Standard Error) for Coverage, Factuality, and Coherence (Task 3) for different settings judged by human experts and GPT-4. The quadratic weighted Cohen’s Kappa with scores indicating the agreement between Expert's judgments and GPT-4 ratings: Coverage $\kappa = 0.50$, Factuality $\kappa = 0.35$, and Coherence $\kappa = 0.04$.}

\label{tab:likeret_combined}
\end{table*}

\paragraph{\textit{Models vs.~Humans}:~Experts prefer model-generated summaries over human-written review abstracts}
\label{sec:model_vs_human}
Results from Task 1:~pairwise comparisons in Table~\ref{tab:comparison} show that model-generated summaries, whether they incorporate hierarchy, are preferred by human domain experts. The review abstracts (human-written summaries) achieved an overall win rate of only 0.2, and are very rarely preferred for having high complexity (0.044) and low clarity (0.089). In contrast, model-generated summaries, even without hierarchical structures, were preferred. Notably, Claude 3 
achieved 0.814 in overall preference, while GPT-4 scored 0.537. Mistral-7B's performance improved significantly in the Recursive-HMDS setting, achieving 0.578 win rate in overall preference, 0.622 in complexity, and 0.622 in clarity, surpassing the Plain-MDS setting (overall: 0.362, complexity: 0.340, clarity: 0.298). This demonstrates the effectiveness of incorporating hierarchies into Mistral-7B.



\paragraph{\textit{Hierarchical vs.~Non-hierarchical MDS}:~Hierarchies can improve Mistral-7B's performance, but may not consistently benefit larger LLMs}

\label{sec:hierarchy}
Figure~\ref{fig:wintie} demonstrates that hierarchical configurations (Recursive-HMDS and HMDS) generally outperform Plain-MDS across most dimensions for Mistral-7B. Mistral-7B shows consistent improvements in overall win rates and other critical aspects, especially in the Recursive-HMDS setting, such as reduced complexity and enhanced clarity and understandability. 
For larger models like GPT-4 and Claude 3, however, the advantages of hierarchical structures are less pronounced. The win rates for these models in hierarchical setups do not consistently exceed those of the 
Plain-MDS setup. While higher win rates are generally observed in hierarchical settings, they are not statistically significant. These results suggest that hierarchical organization in MDS could offer more substantive benefits for smaller LLM like Mistral-7B, but its impact on larger models requires further investigation.

\begin{figure}[t!]
    \centering
    \includegraphics[width=0.48\textwidth]{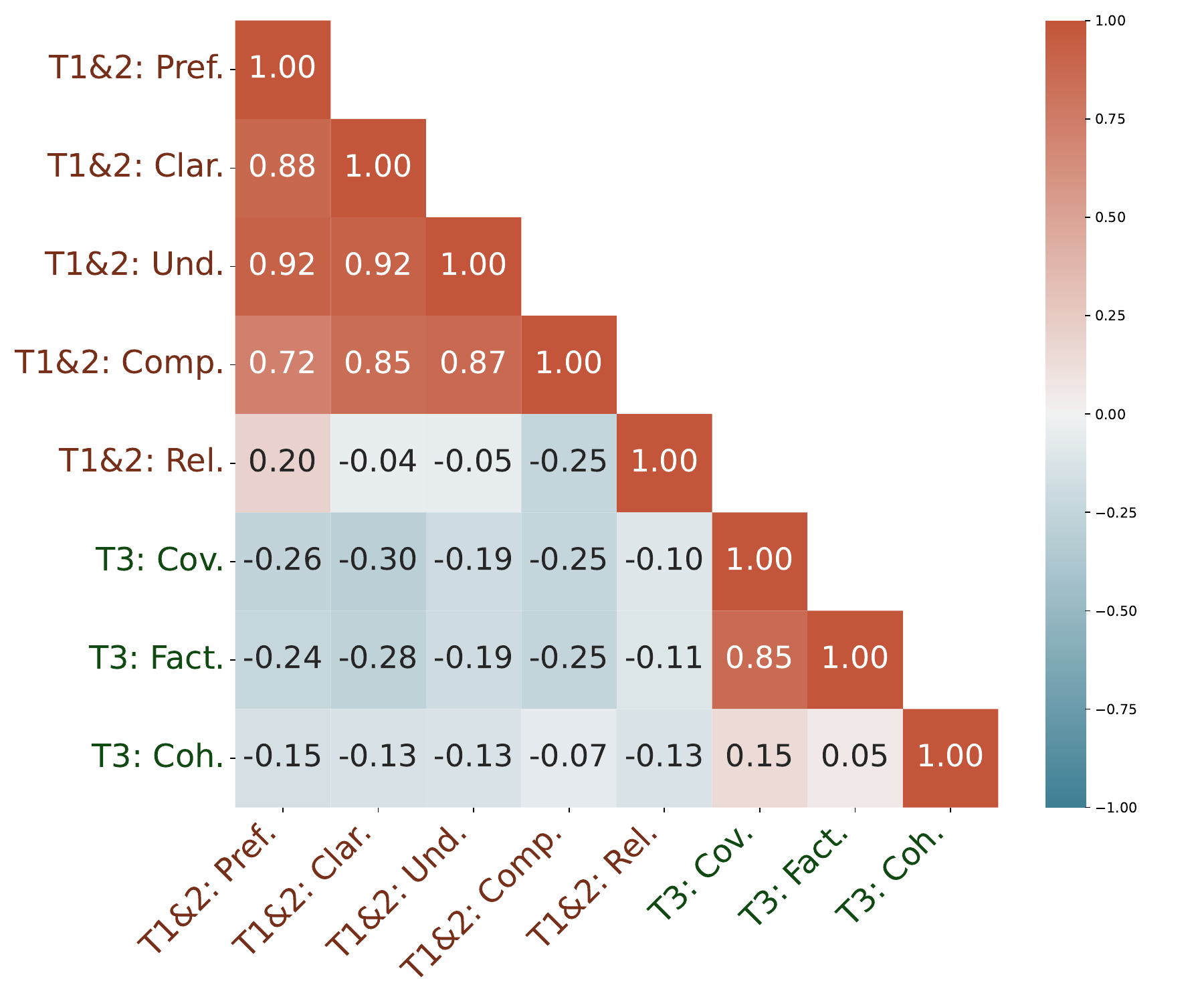}
    \caption{Correlation heat map for Pairwise Comparison (Tasks 1 and 2) and Likert-Scale Assessment (Task 3), with the score mapping method detailed in Appendix~\ref{sec:instance_corr}. 
    }
    \label{fig:heatmap_human_tasks}
\end{figure}

\begin{table}[!ht]
    \centering
    \small 
    \begin{adjustbox}{max width=\linewidth} 
    \begin{tabularx}{\linewidth}{lccccX}
        \midrule
        \textbf{Evaluator} & \multicolumn{5}{c}{\textbf{Percentage Agreement}} \\
        \midrule
            & \textbf{Pref.} & \textbf{Comp.} & \textbf{Rel.} & \textbf{Clar.} & \textbf{Und.}  \\
        \cmidrule{2-6}
        Expert    
        & 0.47 & 0.75 & 0.31 & 0.56 & 0.42 \\
        GPT-4  & 0.56 & 0.67 & 0.36 & 0.56 & 0.50 \\
        GPT-4 (8-shots)     & 0.52 & 0.52 & 0.34 & 0.48 & 0.44 \\
        \midrule \midrule
        & \multicolumn{5}{c}{\textbf{Kappa Agreement}} \\
        \midrule
            & \textbf{Pref.} & \textbf{Comp.} & \textbf{Rel.} & \textbf{Clar.} & \textbf{Und.}  \\
        \cmidrule{2-6}
        Expert      & 0.38 & 0.63 & -0.03 & 0.48 & 0.32 \\
        GPT-4  & 0.5    & 0.62  & 0.12   & 0.5  & 0.45 \\
        GPT-4 (8-shots)     & 0.46  & 0.46  & 0.11   & 0.42   & 0.38 \\
        \midrule
    \end{tabularx}
    \end{adjustbox}
    \caption{Agreement percentages and Kappa scores across evaluators for Tasks 1 and 2. Fleiss' Kappa is computed for four human experts, while Cohen's Kappa is used for GPT-4 evaluations. Pref: Overall Preference, Comp: Complexity, Rel: Relevance, Clar: Clarity, Und: Understandability.}
    \label{tab:human_gpt_aggrement_task1}
\end{table}
\paragraph{Coverage and factuality may not be the primary factors driving expert preference}
\label{sec:coverage_factuality}

The results of the Likert scale assessments in Table \ref{tab:likeret_combined} and the win-tie plot in Figure \ref{fig:wintie} show that while hierarchical summarization with LLM-generated hierarchies lead to reductions in coverage and factuality compared to the Plain MDS baseline (e.g., a decrease of 0.61 in coverage, 0.56 in factuality for Mistral-7B), these changes do not affect expert preference. However, when using expert-curated hierarchies in HMDS, no such discrepancies were observed in both dimensions, highlighting the importance of high-quality hierarchical organization. 
Human experts consistently favored hierarchical summaries, particularly for their enhanced clarity and understandability. This misalignment between traditional metrics and expert preference is further highlighted by Figure \ref{fig:heatmap_human_tasks}, which reveals that factuality and coverage do not correlate with clarity, understandability, or overall preference. This suggests that while coverage and factuality have traditionally been the key focus in evaluating MDS, they might not fully capture the factors driving expert preference in literature reviews. Future work could explore additional factors, as emphasized by \citet{10.1093/pnasnexus/pgae387}, which highlight clarity and understandability as crucial for meeting user needs in effective science communication.

\paragraph{Zero-shot GPT-4-simulated evaluators can reach moderate agreement with human evaluators}
\label{sec:gpt4_evaluator}

We report in Table \ref{tab:human_gpt_aggrement_task1} the inter-rater agreement (IAA) between human experts and GPT-4, with GPT-4 showing the highest agreement on complexity in the zero-shot setting (67\% in percentage agreement and 62\% in Cohen's kappa) and the lowest on relevance (36\% in percentage agreement and 12\% in Cohen's kappa), with moderate alignment on other aspects. This trend mirrors human IAA, where relevance also shows the lowest IAA (67\% in percentage agreement and -3\% in Fleiss' Kappa), closely aligning with the agreement level (4.4\%) reported by \citet{liu-etal-2024-learning} for similar pairwise comparison settings in summarization tasks. These findings reaffirm the hypotheses in \citet{zhang-etal-2024-benchmarking}, highlighting the difficulty of reaching higher IAA on summarization evaluation, especially when comparing summaries of similar quality. 
Simulating annotations in the few-shot setting results in a decline in agreement across all aspects, particularly for complexity (dropping from 67\% to 52\% in percentage agreement and from 62\% to 46\% in Cohen' Kappa), indicating that additional examples do not necessarily enhance alignment with human judgments. 

\begin{figure}[!t]
    \centering
    \includegraphics[width=0.4\textwidth]{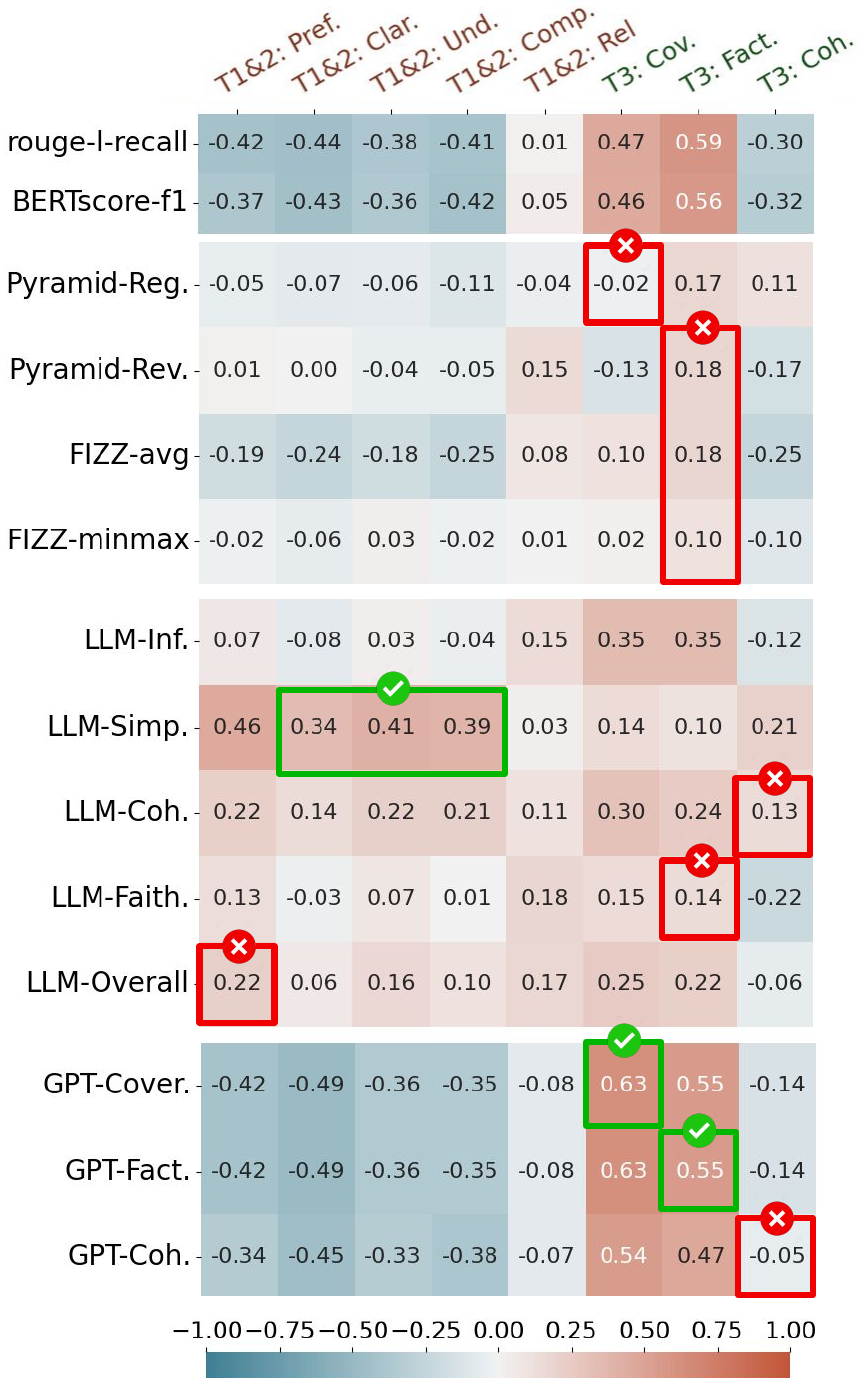}
    \caption{Correlation heat map of human and automatic metrics. Green Boxes highlighted correlations that align with the the metrics. Red boxes indicate misalignment between automated metrics and human judgments.
    } 
    \label{fig:heatmap_human_auto}
\end{figure}

\paragraph{Automated metrics showed limited agreement with human assessments in medical MDS}
\label{sec:metrics_corr_human}
Prior work highlights that human experts often disagree —especially when the differences in summaries are subtle or hard to discern \cite{liu-etal-2024-learning, zhang-etal-2024-benchmarking}. We further investigate how this variation in human judgment might influence the correlation between automated metrics and human assessments. 
The correlation heatmap (Figure~\ref{fig:heatmap_human_auto}) of human judgments 
indicates that GPT-4 agrees well with human assessments in Coverage (0.63) and Factuality (0.55), where summaries are evaluated with respect to paper abstracts. This suggests that GPT-4's outputs generally concur with human judgments when the underlying information is provided \cite{Doostmohammadi2024HowRA}. There is, however, weak to moderate correlation with human subjective preferences for simplicity (0.46), clarity (0.34), understandability (0.41), and complexity (0.39). 
Lastly, BertScore and ROUGE achieve moderate correlation 
(0.56 and 0.59) with factuality, showing their effectiveness in evaluating semantic similarity as a proxy for assessing factual accuracy. Conversely, 
some metrics designed to capture factuality (Pyramid-reversed, FIZZ, and LLM-faithfulness) and coherence by GPT-4 simulation show weak correlations to human judgments. 
These results confirm that automated and model-based metrics are insufficient replacements for expert evaluation, as suggested by prior work \citep{wang2023automated}.

\section{Discussion \& Conclusion}
\label{sec:discussion}
Our study demonstrates the benefits of hierarchical organization 
for improving medical MDS. Models with hierarchical organization of inputs consistently outperform Plain MDS in expert perceived clarity, understandability, and overall preference, showcasing the value of structured summarization in MDS. We also find that although existing  automated metrics fail to capture human judgments accurately, GPT-4 evaluation aligns well with human judgments on factuality and coverage, but exhibits discrepancies in subjective dimensions like relevance. 
Future work can explore integrating hierarchical summaries into interactive systems to enhance user engagement. Additionally, examining annotations in practical settings is essential for understanding the utility of summaries, 
and refining evaluation frameworks to more effectively capture both summary quality and domain-specific user preferences.

\section*{Limitations}
While the use of human experts for evaluation offered valuable insights, it may introduce potential biases, as expert judgments are inherently subjective and can vary depending on individual interpretation. To mitigate this, we engaged six domain experts to complete the entire annotation task, but the possibility of variability in expert assessments remains a potential issue.

Second, although we compared a range of models, including GPT-4, Claude, and Mistral, future research should explore a wider variety of open-source models, particularly for smaller-model applications, to assess whether similar patterns hold across different models and explore how these models, often more accessible and cost-effective, can be leveraged in practical scenarios.


Finally, while our study explores hierarchical organization in medical MDS, its applicability to other domains, particularly non-medical fields, remains an open question. The impact of hierarchical methods on summaries in other specialized domains may vary due to cultural differences and domain-specific needs. More high-quality datasets and evaluation procedures are needed to study the generalization of our findings in these broader application settings.
\section*{Ethical Statement}
We identify no immediate ethical concerns with our research and development. The data for our corpus was sourced from publicly accessible web resources. Our six annotators were compensated at or above the minimum wage as per the standards of their respective regions, with a commitment to a maximum of 20 working hours. Additionally, we believe the summaries generated by our models are unlikely to be misused for ethically questionable purposes.
\bibliography{custom}

\appendix
\label{sec:appendix}

\section{Automated Metrics}
\label{sec:app_metrics}

We provide additional details for metrics and metric implementation below:

\paragraph{Model-Based Metrics}
\begin{itemize}[leftmargin=10pt, itemsep=2pt, topsep=2pt]

 \item{BERT-Score} \cite{DBLP:journals/corr/abs-1904-09675}: A BERT-based similarity metric that evaluates the semantic similarity between the generated text and the source text.

\item{Pyramid Score} \cite{zhang2021finding} uses GPT-4 (gpt-4-turbo-2024-04-09) to generate Summary Content Units (SCUs) \cite{nawrath-etal-2024-role} from the source documents and use entailment model to calculate how much of the SCUs are entailed by the generated summary, enabling a structured evaluation of content coverage. In addition, we report \textit{Reversed-Pyramid}, which generates SCUs from the generated summary and calculates the entailment score of the SCUs in the source documents. This metric provides insight into the extent of hallucination in the summaries generated. A higher Pyramid score indicates that more information units from the source text are covered in the generated summary, whereas a higher Reversed-Pyramid score indicates that more information units from the generated text are grounded in the source text.

 \item{FIZZ} \cite{yang-etal-2024-fizz}: A factual consistency metric that measures the entailment between atomic facts in the generated summaries and the source document. Besides the min-max score defined in the original paper, We report two variants of FIZZ: (i) min-max: The Min-Max score computed on the entailment scores of atomic facts compared to every sentence in the document. (ii) average: The average score of the entailment scores of atomic facts compared to every sentence in the document.
    

\end{itemize}
    
\paragraph{LLM Prompt-based Metrics} prompts LLMs for human-like summarization evaluation.  We adopt the prompt templates from \citet{guo-etal-2024-appls} and \citet{gao2023human} to prompt GPT-4 (gpt-4-turbo-2024-04-09) to evaluate each summary in terms of informativeness, simplification, coherence, and faithfulness, and to provide an overall quality score.



\section{Impact of Condensed Claims on Summary Quality}
\label{sec:condense}
Summarizing multiple paper abstracts often results in reduced automatic metrics like BERTScore, ROUGE, Pyramid, and FIZZ for smaller models like Mistral as shown in Table \ref{tab:autoresults}. However, condensing source documents into a single sentence that captures the main claim of each paper can significantly improve performance across these facets. In contrast, large language models (LLMs) are better equipped to process long-format input documents, and their performance remains largely unaffected by the length of the input documents.

\section{Mapping Pairwise Evaluation onto Instance-Level Scores}
\label{sec:instance_corr}
To enable comparison of automated metrics and human pairwise preferences, we convert the binary preferences (i.e., Summary A vs. Summary B) of human experts into an overall ranking for all settings and summaries. We develop the following approach for each measure: for each comparison made by an expert, we add one point to the preferred setting and subtract one point from the un-preferred setting; if the annotator chooses ``About the Same,'' we do not add any points to either setting. We then normalize these points by the total number of comparisons made for each instance in each setting. 
\begin{figure*}[ht]
    \centering
    \includegraphics[width=\textwidth]{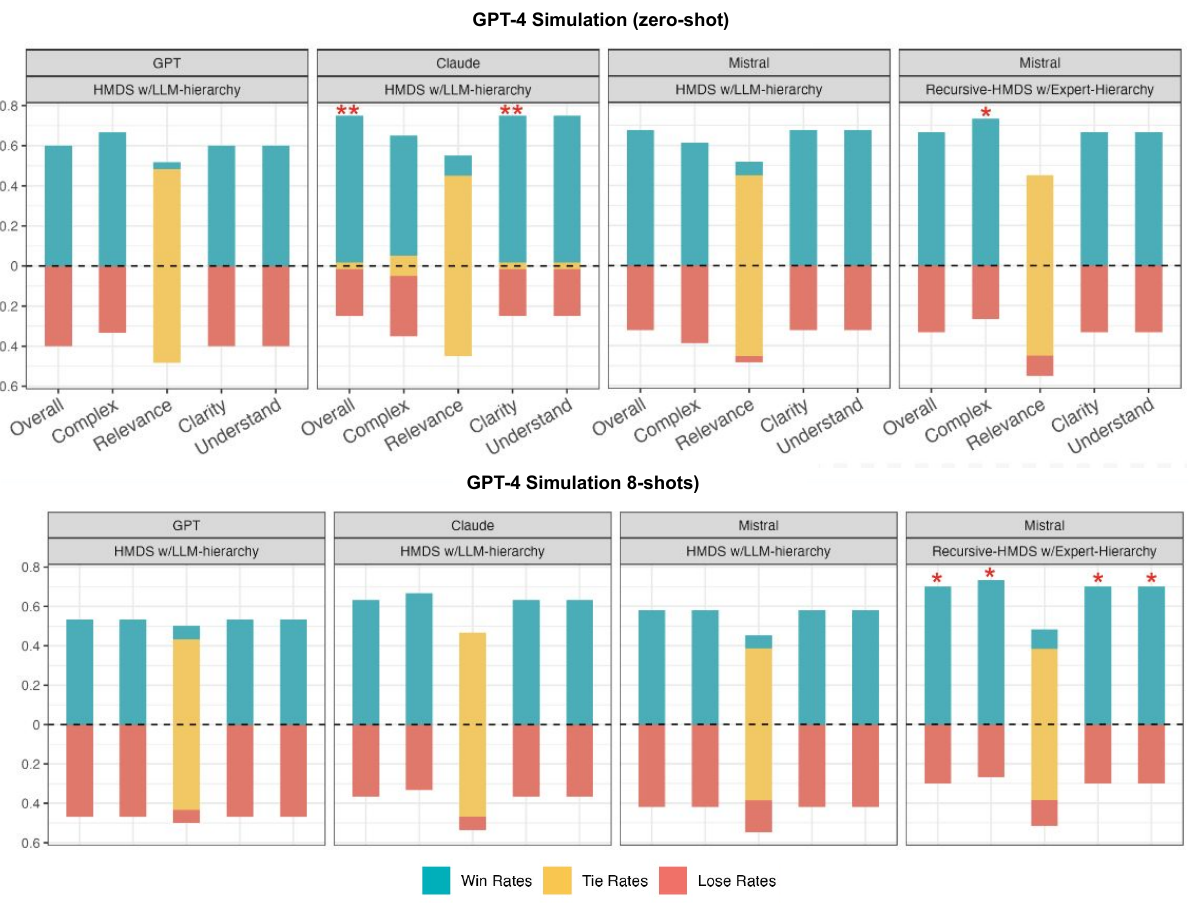}
    \caption{GPT-4-simulation (zero-shot and 8-shots) pairwise evaluations of LLM-generated summaries of hierarchical setups compared to the baseline (Plain MDS). We report 1-Complexity to ensure alignment in the direction of all dimensions. The red star (\textcolor{red}{$\star$}) indicates the significance levels, with \textcolor{red}{$\star$$\star$} (p < 0.01) and \textcolor{red}{$\star$} (p < 0.05), as determined by McNemar's test (within-subject Chi-square test) based on the counts of wins and ties of each pairs.}
    \label{fig:gpt-8shots}
\end{figure*}

\begin{table*}[ht]
\centering
\caption{Automated metric scores}
\begin{adjustbox}{width=\textwidth}
\begin{tabular}{ll|l|ccc|c|cc|cc|ccc}
\toprule
\multirow{2}{*}{\textbf{Setting}} & & \multirow{2}{*}{\textbf{Model}} & \multicolumn{3}{c|}{\textbf{BERTScore}} & \multirow{2}{*}{\textbf{rouge-l-recall}} & \multicolumn{2}{c|}{\textbf{Pyramid}} & \multicolumn{2}{c|}{\textbf{FIZZ}} & \multirow{2}{*}{\textbf{Sent. len}} & \multirow{2}{*}{\textbf{Para. len}} & \multirow{2}{*}{\textbf{Total len.}} \\

& & &  \textbf{p} & \textbf{r} & \textbf{f1} & & \textbf{Regular} & \textbf{Reversed} & \textbf{avg} & \textbf{minmax} & & & \\
\midrule
\multicolumn{2}{l|}{Review Abstract} & - & 0.56 & 0.55 & 0.55 & 0.19 & 0.22 & 0.30 & 0.29 & 0.00 & 30.06 & 12.62 & 378.63 \\

\midrule

\multirow{6}{*}{Plain MDS} 
& \multirow{3}{*}{Abstract} 
    & GPT     & 0.64 & 0.62 & 0.63 & 0.29 & 0.15 & 0.40 & 0.56 & 0.01 & 29.22 & 12.49 & 365.01 \\
&   & Claude   & 0.65 & 0.57 & 0.61 & 0.19 & 0.10 & 0.38 & 0.55 & 0.01 & 22.69 & 7.70  & 174.66 \\
&   & Mistral  & 0.65 & 0.61 & 0.63 & 0.25 & 0.14 & 0.40 & 0.54 & 0.01 & 27.31 & 11.94 & 311.17 \\
\cmidrule(lr){2-14}
& \multirow{3}{*}{Claim Statement} 
    & GPT     & 0.67 & 0.63 & 0.65 & 0.28 & 0.14 & 0.40 & 0.67 & 0.05 & 28.63 & 9.28  & 275.55 \\
&   & Claude   & 0.67 & 0.61 & 0.65 & 0.27 & 0.12 & 0.45 & 0.69 & 0.07 & 24.93 & 7.53  & 187.78 \\
&   & Mistral  & 0.76 & \textbf{0.71} & \textbf{0.74} & \textbf{0.49} & 0.22 & 0.49 & 0.82 & 0.13 & 24.20 & 11.22 & 514.80 \\

\midrule
\multirow{6}{*}{Hierarchical-MDS (HMDS)} 
& \multirow{3}{*}{w/ LLM-Hierarchy} 
    & GPT     & 0.70 & 0.65 & 0.67 & 0.31 & 0.15 & 0.49 & 0.74 & 0.08 & 29.75 & 10.11  & 300.76 \\
&   & Claude   & 0.69 & 0.59 & 0.64 & 0.23 & 0.13 & 0.46 & 0.70 & 0.07 & 24.34 & 8.02   & 195.00 \\
&   & Mistral  & 0.75 & 0.68 & 0.71 & 0.40 & 0.20 & 0.50 & 0.81 & \textbf{0.15} & 28.83 & 12.24 & 353.04 \\
\cmidrule(lr){2-14}
& \multirow{3}{*}{w/ Expert-Hierarchy} 
    & GPT     & 0.69 & 0.65 & 0.67 & 0.32 & 0.13 & 0.47 & 0.71 & 0.10 & 29.37 & 10.96  & 300.10 \\
&   & Claude   & 0.69 & 0.61 & 0.64 & 0.25 & 0.13 & 0.46 & 0.68 & 0.09 & 24.90 & 8.01  & 199.29 \\
&   & Mistral  & \textbf{0.77} & \textbf{0.71} & \textbf{0.74} & 0.45 & \textbf{0.25} & \textbf{0.52} & \textbf{0.85} & 0.13 & 28.53  & 12.55 & 354.03 \\

\midrule
\multirow{6}{*}{HMDS} 
& \multirow{3}{*}{w/ LLM-Hierarchy} 
    & GPT     & 0.63 & 0.58 & 0.61 & 0.22 & 0.12 & 0.38 & 0.54 & 0.02 & 28.80   & 8.16  & 234.92 \\
&   & Claude   & 0.65 & 0.56 & 0.60 & 0.19 & 0.09 & 0.32 & 0.53 & 0.01 & 27.31 & 5.82  & 158.93 \\
&   & Mistral  & 0.68 & 0.63 & 0.65 & 0.29 & 0.17 & 0.50 & 0.62 & 0.04 & 27.16 & 10.77 & 259.05 \\
\cmidrule(lr){2-14}
& \multirow{3}{*}{w/ Expert-Hierarchy} 
    & GPT     & 0.63 & 0.58 & 0.61 & 0.22 & 0.10 & 0.40 & 0.56 & 0.02 & 29.38 & 8.71  & 229.40 \\
&   & Claude   & 0.66 & 0.57 & 0.61 & 0.22 & 0.10 & 0.41 & 0.58 & 0.02 & 27.49 & 5.70  & 186.50 \\
&   & Mistral  & 0.70 & 0.63 & 0.66 & 0.29 & 0.14 & 0.45 & 0.68 & 0.03 & 25.35 & 9.59  & 252.43 \\

\bottomrule
\end{tabular}
\end{adjustbox}
\label{tab:autoresults}
\end{table*}

\begin{figure*}[ht]
    \centering
    \includegraphics[width=\textwidth]{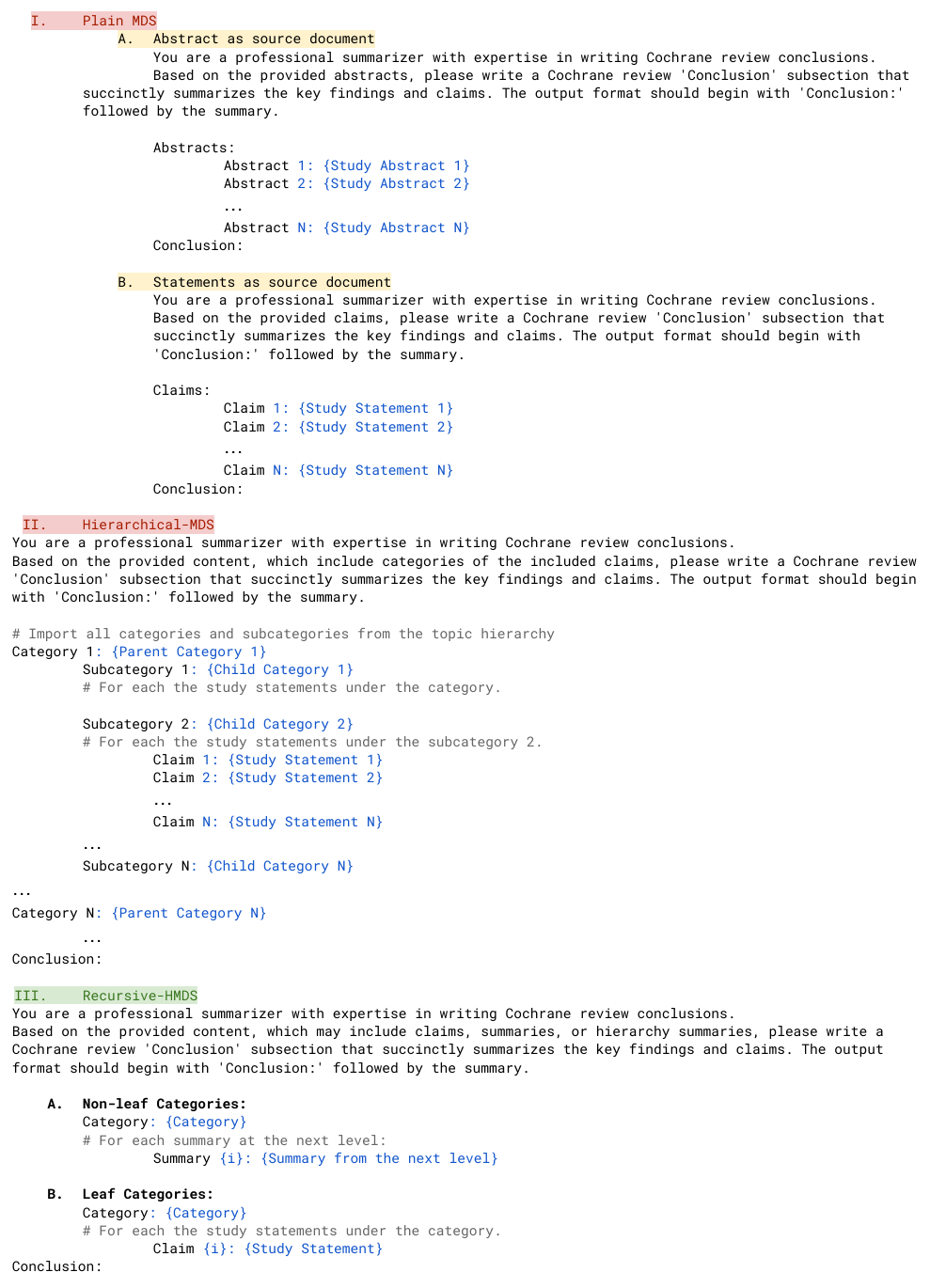}
    \caption{Prompt used for Summaries Generation}
    \label{fig:prompt}
\end{figure*}

\begin{figure*}[ht]
    \centering
    \includegraphics[width=1.15\textwidth]{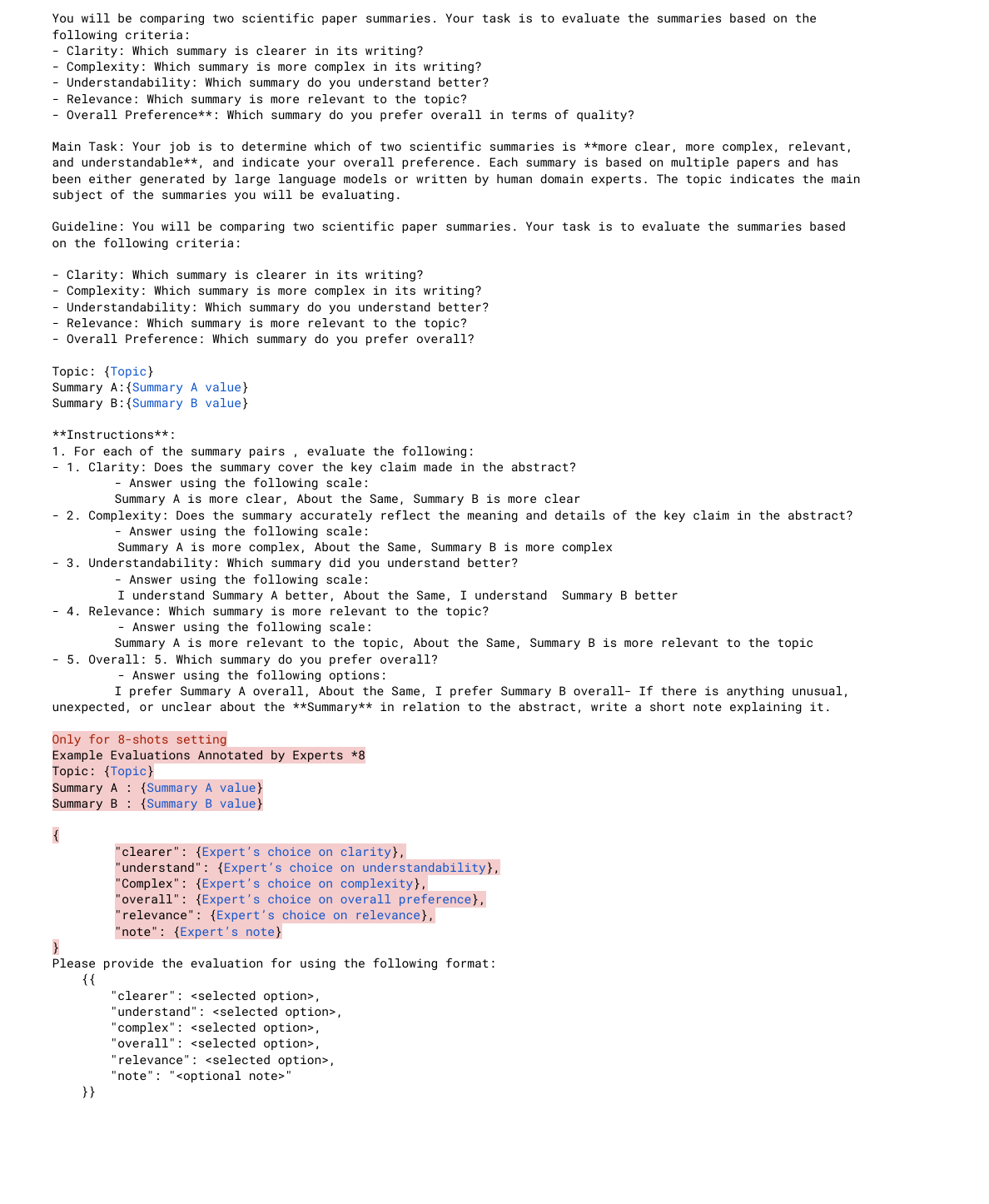}
    \caption{Prompt used for GPT-4 Simulated Evaluation in Task 1 and Task 2}
    \label{fig:prompt_task1}
\end{figure*}

\begin{figure*}[ht]
    \centering
    \includegraphics[width=1.15\textwidth]{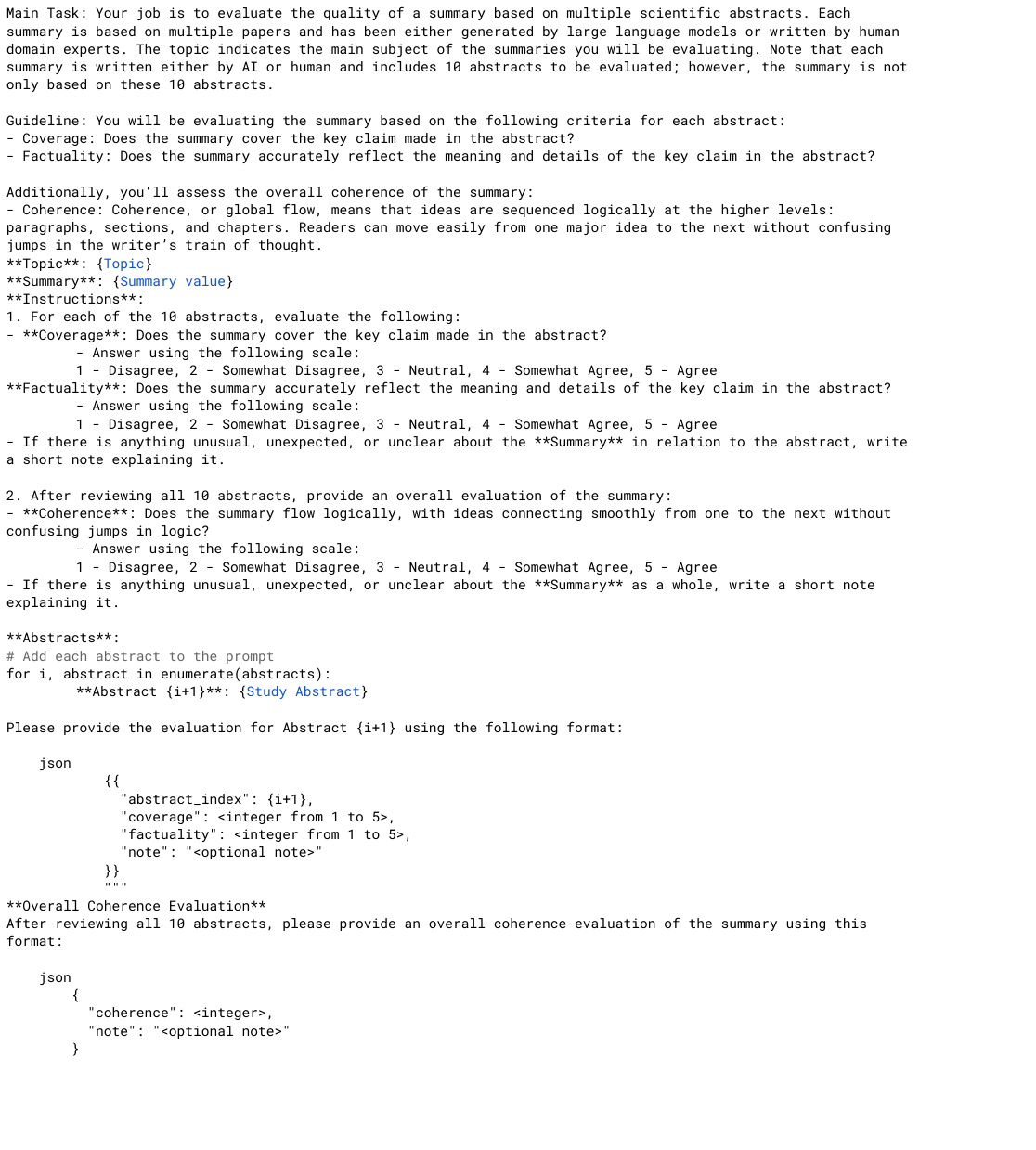}
    \caption{Prompt used for GPT-4 Simulated Evaluation in Task 3}
    \label{fig:prompt_task2}
\end{figure*}


\begin{figure*}[ht]
    \centering
    \includegraphics[width=\textwidth]{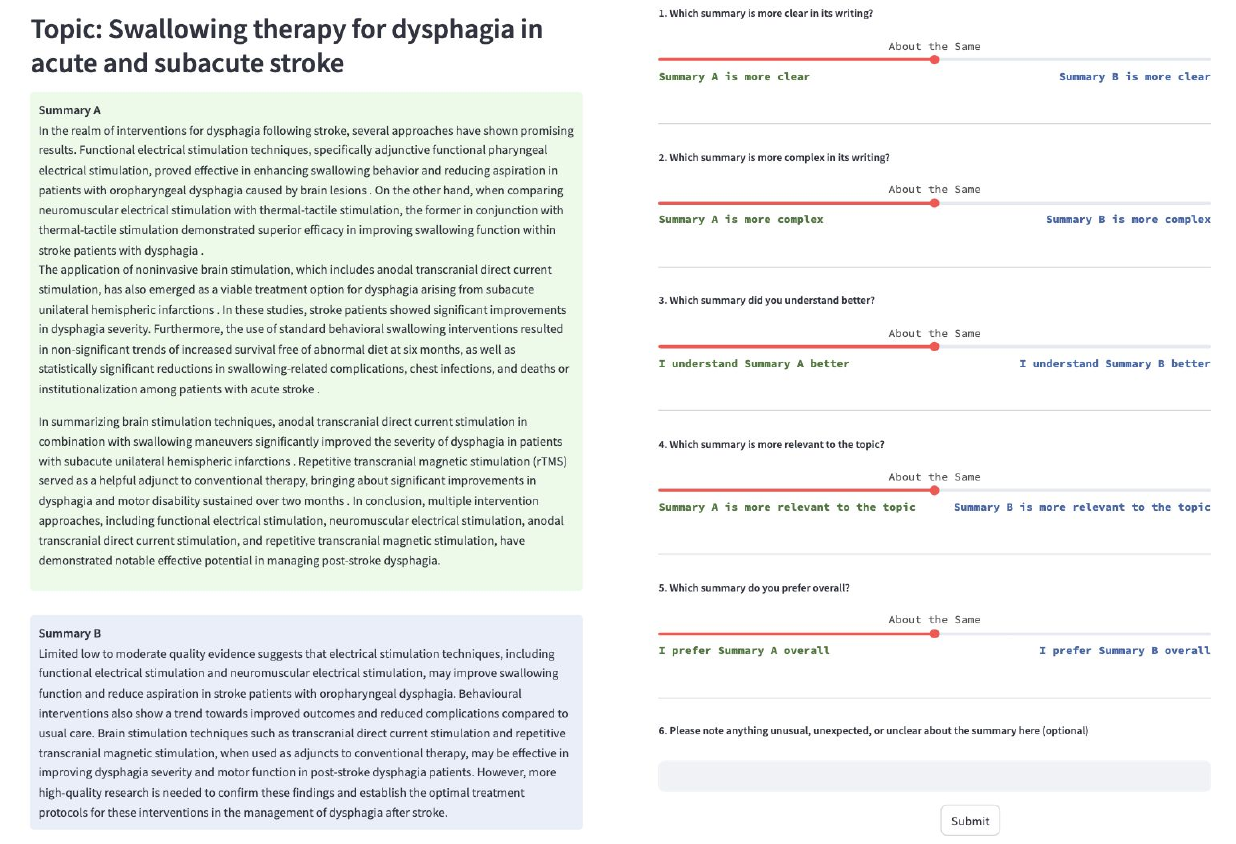} 
    \caption{An example of human evaluation Task 1 and Task 2 - pairwise comparison}
    \label{fig:eval}
\end{figure*}

\begin{figure*}[ht]
    \centering
    \includegraphics[width=\textwidth]{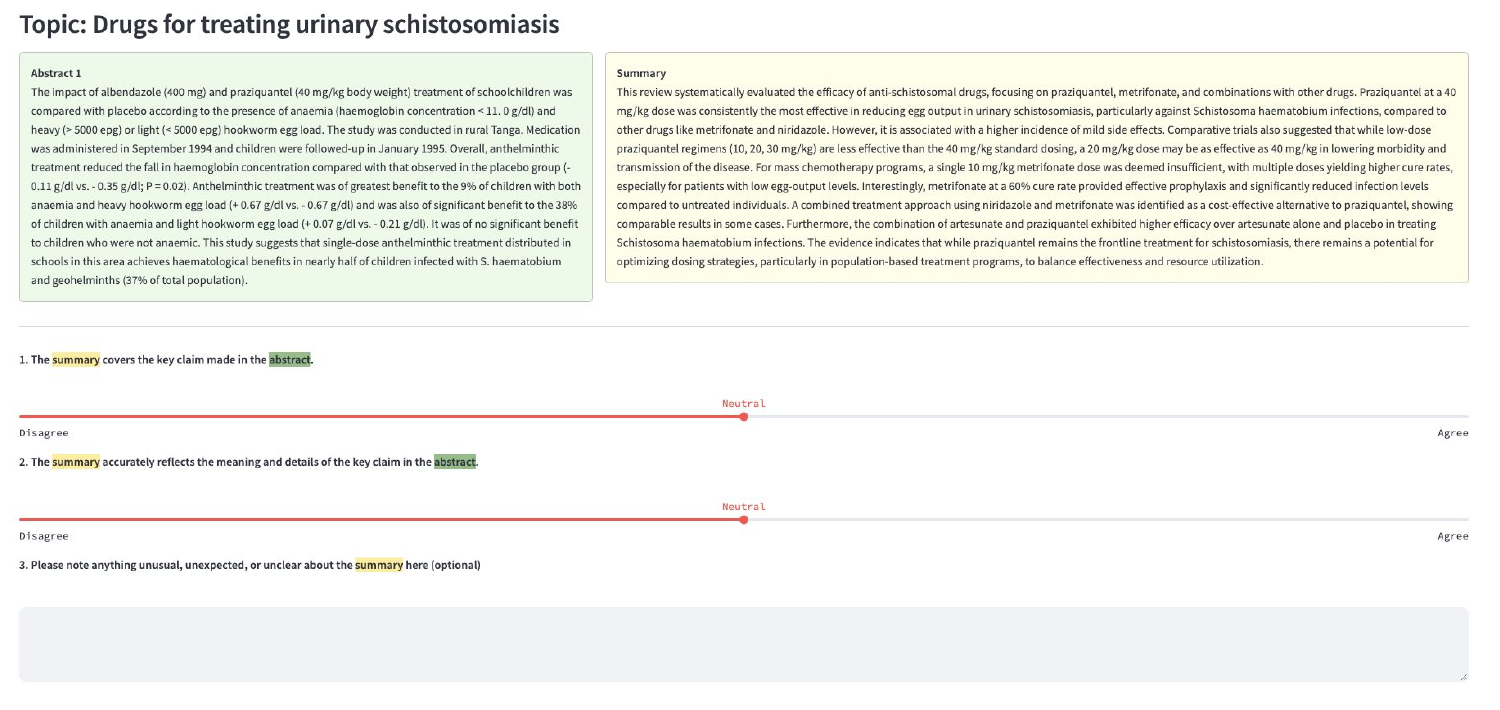}
    \caption{An example of human evaluation Task 3 - coverage and factuality}
    \label{fig:eval-2}
\end{figure*}

\begin{figure*}[ht]
    \centering
    \includegraphics[width=\textwidth]{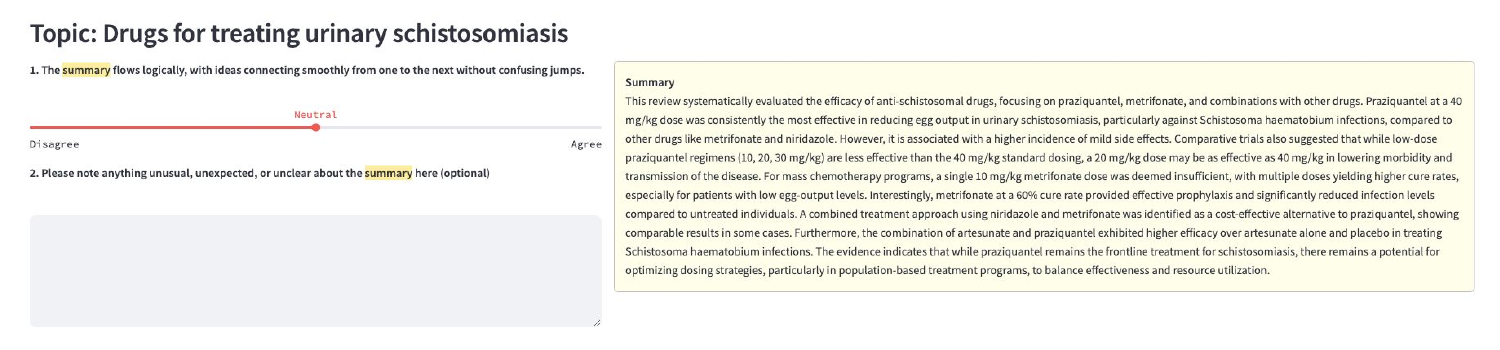}
    \caption{An example of human evaluation Task 3 - coherence}
    \label{fig:eval-3}
\end{figure*}

\end{document}